\def\year{2017}\relax
\documentclass[letterpaper]{article}
\usepackage{aaai17}
\usepackage{times}
\usepackage{helvet}
\usepackage{courier}
\usepackage{amsmath}
\usepackage{amssymb}
\usepackage{graphicx}
\usepackage{url}
\usepackage{algorithm}
\usepackage{algorithmic}
\begin{document}
%
\title{Partial Least Squares Regression on Riemannian Manifolds \\and Its Application in Classifications}

\author{Haoran Chen$^{1}$, Yanfeng Sun$^{1}$, Junbin Gao$^{2}$, Yongli Hu$^{1}$ and Baocai Yin$^{1}$\\[1mm]
$^{1}$Beijing Key Laboratory of Multimedia and Intelligent Software Technology\\
College of Metropolitan Transportation, Beijing University of Technology,
Beijing, 100124, China\\
hr\_Chen@emails.bjut.edu.cn, \{yfsun,huyongli,ybc\}@bjut.edu.cn\\
\and
$^2$ Discipline of Business Analytics,The University of Sydney Business School\\
The University of Sydney,
 Camperdown NSW 2006, Australia\\
junbin.gao@sydney.edu.au
}

\maketitle
\begin{abstract}
Partial least squares regression (PLSR) has been a popular technique to explore the linear relationship between two datasets. However, most of algorithm implementations of PLSR  may only achieve a suboptimal solution through an optimization on the Euclidean space. In this paper, we propose several novel PLSR models on Riemannian manifolds and develop optimization algorithms based on Riemannian geometry of manifolds. This algorithm can calculate all the factors of PLSR globally to avoid suboptimal solutions. In a number of experiments, we have demonstrated the benefits of applying the proposed model and algorithm to a variety of learning tasks in pattern recognition and object classification.

\end{abstract}

\section{Introduction}\label{intro}

Partial least squares regression (PLSR) is a statistical method for modeling a linear relationship between two data sets, which may be two different descriptions of an object. Instead of finding hyperplanes of maximum variance of the original datasets, it finds the maximum  degree of linear association  between two latent components which are the projection of two original data sets to a new space, and based on those latent components, regresses the loading matrices of the two original datasets, respectively. Compared with the multiple linear regression (MLR)\cite{AikenWestPitts2003} and principal component regression (PCR) \cite{Kendall1957,Jolliffe1982}, PLSR has also been proved to be not only useful for high-dimensional data  \cite{HuangPanGrindleHan2005,BoulesteixStrimmer2007}, but also to be a good alternative because it is more robust and adaptable \cite{WoldRuheWoldDunn1984}. Robust means that the model parameters do not change very much when new training samples are taken from the same total population. Thus PLSR  has wide applications in several areas of scientific research \cite{LitonHelenDasIslamKarim2015,HaoThelenGao2016,Worsley1997,Hulland1999,LobaughWestMcIntosh2001} since the 1960s.

There exist many  forms of PLSR, such as NIPALS (the nonlinear iterative partial least squares)\cite{Wold1975}, PLS1 (one of the  data sets consists of a single variable)\cite{Hoskuldsson1988} and PLS2 (both data sets are multidimensional) where a linear inner relation between the projection vectors exists, PLS-SB \cite{Wegelin2000,RosipalKramer2006}, where the extracted   projection matrices are in general not mutually orthogonal, statistically inspired modification of PLS (SIMPLS) \cite{Jong1993}, which calculates the PLSR factors directly as linear combinations of the original data sets, Kernel PLSR \cite{Rosipal2003} applied in a reproducing kernel Hilbert space, and Sparse PLSR \cite{ChunKeles2010} to achieve factors selection by producing sparse linear combinations of the original data sets.

However, it is difficult to directly solve for projection matrices with orthogonality as a whole in Euclidean spaces. To the best of our knowledge, all the existing algorithms greedily proceed through a sequence of low-dimensional subspaces: the first dimension is chosen to optimize the PLSR objective, e.g., maximizing the covariance between the projected data sets, and then subsequent dimensions are chosen to optimize the objective on a residual or reduced data sets. In some sense, this  can be actually fruitful but limited, often resulting in ad hoc or suboptimal solutions. To overcome the shortcoming, we are devoted to proposing  several novel models and algorithms to solve PLSR problems under the framework of Riemannian manifold optimisation \cite{AbsilMahonySepulchre2008}. For the optimisation problems from PLSR, the orthogonality constraint can be easily eliminated in Stiefel/Grassmann manifolds with the possibility of  solving the factors of PLSR as a whole and  being steadily convergent at global optimum.

In general, Riemannian optimization is directly based on the curved manifold geometry such as Stiefel/Grassmann manifolds,  benefiting from a lower complexity and better numerical properties. The geometrical framework of  Stiefel and Grassmann manifolds were proposed in \cite{EdelmanAriasSmith1998}. Stiefel manifold was successfully applied in neural networks \cite{NishimoriAkaho2005} and linear dimensionality reduction \cite{CunninghamGhahramani2014}.  Meanwhile, Grassmann manifold has been studied in two major fields, data analysis such as video stream analysis \cite{HeBalzanoSzlam2012}, clustering subspaces into classes of subspaces \cite{WangHuGaoSunYin2014,WangHuGaoSunYin2016a},  and parameter analysis such as an unifying view on the subspace-based learning method \cite{HammLee2008},  and optimization over the Grassmann manifold \cite{MishraSepulchre2014,MishraMeyerBonnabelSepulchre2014}. According to \cite{EdelmanAriasSmith1998,AbsilMahonySepulchre2004}, the generalized Stiefel manifold is endowed with a scaled metric by making it a Riemannian submanifold based on Stiefel manifold, which is more flexible to the constraints of the optimization  raised from the generalised PLSR. Generalized Grassmann manifold  is generated by the Generalized Stiefel manifold, and each point on this manifold is a collection of ``scaled'' vector subspaces of dimension $p$ embedded in $\mathbb R^n$. Another important matrix manifold is the oblique manifold which is  a product of spheres. Absil $\emph{et al.}$ \cite{AbsilGallivan2006} investigate the geometry of this manifold and show how independent component analysis can be cast on this manifold as non-orthogonal joint diagonalization.

Some conceptual algorithms and its convergence analysis based on ideas of Riemannian manifolds, and the efficient numerical implementation \cite{AbsilMahonySepulchre2008} have been developed recently. This has paved the way for one to investigate overall algorithms to solve PLSR problems based on optimization algorithms on Riemannian manifolds. Particularly, Mishra $\emph{et al.}$ \cite{BoumalMishraAbsil2014} have developed a useful MATLAB toolbox ManOpt (Manifold Optimization) \url{http://www.manopt.org/} which can be perfectly adopted in this research to test the algorithms to be developed.

The contributions of this paper are:
\begin{enumerate}
   \item We establish several novel PLSR models on Riemannian manifolds and give some matrices representations of relate optimization ingredients;
   \item We give new algorithms  for the proposed PLSR model on Riemannian manifolds, which are able to calculate all the factors as a whole so as to obtain optimal solutions.
\end{enumerate}


\section{Notations and Preliminaries}\label{nota}
This section will briefly describe some notations and concepts that will be used throughout the paper.
\subsection{Notations}
We denote matrices by boldface capital letters, \emph{e.g.}, $\mathbf{A}$, vectors by boldface lowercase letters, \emph{e.g.}, $\mathbf{a}$, and scalars by letters, \emph{e.g.}, $a$.   The superscript $T$ denotes the transpose of a vector/matrix. $\text{diag}(\mathbf A)$ denotes the diagonal matrix with elements from the diagonal of $\mathbf A$.  $\mathbf B \succ 0$ means that $\mathbf B$ is a positive definite matrix. The SVD decomposition of a matrix $\mathbf A\in\mathbb{R}^{m\times n}$ is denoted by $\mathbf A = \mathbf U \boldsymbol{\Sigma}\mathbf V^T$, while the eigendecomposition of a diagonable square matrix $\mathbf A\in\mathbb{R}^{n\times n}$ is denoted by $\mathbf A = \mathbf E_v\mathbf E_{\lambda}\mathbf E_v^{-1}$.

The set of all $c$-order orthogonal matrices is denoted by
\[
\mathcal{O}(c) = \{\mathbf U\in\mathbb R^{c\times c}| \mathbf U^T\mathbf U =\mathbf U\mathbf U^T= \mathbf I_c\},
\]
also called orthogonal group of order $c$. The Stiefel manifold is the set of all the matrices whose columns are orthogonal, denoted by
\begin{equation}\label{st}
St(p,c) = \{\mathbf W\in\mathbb{R}^{p\times c}| \mathbf W^T\mathbf W = \mathbf I_c\}.
\end{equation}

Given a Stiefel manifold $St(p,c)$, the related Grassmann manifold $Gr(p,c)$ can be formed as the quotient space of $St(p,c)$ under the equivalent relation defined by the orthogonal group $\mathcal{O}(c)$, i.e.
\begin{equation}\label{gr}
G(p,c) = St(p,c)/\mathcal{O}(c).
\end{equation}

Two Stiefel points $\mathbf W_1, \mathbf W_2 \in St(p,c)$ are equivalent to each other, if there exists an $\mathbf O\in\mathcal{O}(c)$ such that $\mathbf W_1  = \mathbf W_2 \mathbf O$. We use $[\mathbf W]\in Gr(p,c)$ to denote the equivalent class for a given $\mathbf W\in St(p.c)$, and $\mathbf W$ is called a representation of the Grassmann point $[\mathbf W]$. More intuitively, Grassmann manifold is the set of all $c$-dimensional subspaces in $\mathbb{R}^p$.

In this paper, we are also interested in the so-called generalized Stiefel manifold which is defined under the $\mathbf B$-orthogonality
\begin{equation}\label{gst}
GSt(p,c; \mathbf B) = \{\mathbf W\in\mathbb{R}^{p\times c} | \mathbf W^T\mathbf B\mathbf W = \mathbf I_p\},
\end{equation}
where $\mathbf B\in\mathbb{R}^{c\times c}$ is a given positive definite matrix.  And similarly the generalized Grassmann manifold is defined by
\begin{equation}\label{ggr}
GGr(p,c;\mathbf B) = GSt(p,c; \mathbf B)/\mathcal{O}(c).
\end{equation}

If we relax the orthogonal constraints but retain unit constraint, we have the so-called Oblique manifold which consists of all the $p\times c$ matrices whose columns are unit vectors. That is
\begin{equation}\label{obl}
Ob(p,c)=\{\mathbf W\in\mathbb R^{p\times c}, \text{diag}(\mathbf W^T\mathbf W) =\mathbf I_c \}.
\end{equation}

\subsection{Partial Least Squares Regression (PLSR)}
Let $\mathbf{x}_i=[x_{i1},x_{i2},\ldots,x_{ip}]^T\in\mathbb{R}^p$, $(i=1,2,\ldots,n)$ are $n$ observation samples and $\mathbf{y}_i=[y_{i1},y_{i2},\ldots,y_{iq}]^T\in\mathbb{R}^q$, $(i=1,2,\ldots,n)$ are $n$ response data. Then $\mathbf{X}=[\mathbf{x}_1,\mathbf{x}_2,...,\mathbf{x}_n]^T\in\mathbb{R}^{n\times p}$, $\mathbf{Y}=[\mathbf{y}_1,\mathbf{y}_2,...,\mathbf{y}_n]^T\in\mathbb{R}^{n\times q}$.

Suppose there exists a linear regression relation
\begin{equation}\label{ob}
\mathbf Y = \mathbf X\mathbf R+\mathbf E,
\end{equation}
where $\mathbf R$ is the regression coefficient and $\mathbf E$ is the residual matrix. PLSR is usually an  effective approach to dealing  with the case of  $n<p$ when the classical linear regression fails since the $p\times p$ covariance matrix $\mathbf X^T\mathbf X$ is singular.

In order to obtain $\mathbf R$, PLSR generally decomposes  datasets $\mathbf X$ and $\mathbf Y$ ($\mathbf X$ and $\mathbf Y$ are preprocessed to be zero-mean data) into the following form
\begin{equation}\label{pls1}
\begin{split}
&\mathbf X_{n\times p}  = \mathbf t_{n\times 1}\mathbf p_{p\times 1}^T+\mathbf E_{n\times p}\\
&\mathbf Y_{n\times q} = \mathbf u_{n\times 1}\mathbf q_{q\times 1}^T+\mathbf F_{n\times q}
\end{split}
\end{equation}
where $\mathbf t$ and $\mathbf u$ are  vectors giving the latent components for the $n$ observations, $\mathbf p$ and  $\mathbf q$ represent loading vectors. $\mathbf E$ and  $\mathbf F$ are residual matrices.

PLSR searches the latent  components $\mathbf t= \mathbf X\mathbf w$ and $\mathbf u= \mathbf Y\mathbf g$ such that the squared covariance between them is maximized, where the projection vectors $\mathbf w$ and $\mathbf g$ satisfy the constraints $\mathbf w^T\mathbf w = 1$ and $\mathbf g^T\mathbf g = 1$, respectively.  The solution is given by
\begin{equation}\label{tu}
\begin{split}
\max_{\|\mathbf w\|=\|\mathbf g\|=1}[\text{cov}(\mathbf t, \mathbf u)]^2 &= 
 \max_{\|\mathbf w\|=\|\mathbf g\|=1} (\mathbf w^T\mathbf X^T\mathbf Y\mathbf g)^2.
\end{split}
\end{equation}

It can be shown that the projection  vector $\mathbf w$ corresponds to the first eigenvector of $\mathbf X^T\mathbf Y\mathbf Y^T\mathbf X$ \cite{Hoskuldsson1988,RosipalKramer2006}
and the optimal solution $\mathbf w$ of
\begin{equation}\label{ww}
\max_{\|\mathbf w\|=1} \mathbf w^T\mathbf X^T\mathbf Y\mathbf Y^T\mathbf X\mathbf w
\end{equation}
is  also the first eigenvector of $\mathbf X^T\mathbf Y\mathbf Y^T\mathbf X$. Thus both objectives \eqref{tu} and \eqref{ww} have the same solution on $\mathbf w$.

We can also obtain $\mathbf g$ while swapping the position of $\mathbf X$ and $\mathbf Y$. After obtaining the projection vectors $\mathbf w$ and $\mathbf g$, the latent vectors $\mathbf t = \mathbf X\mathbf w$ and $\mathbf u = \mathbf Y\mathbf g$ are also acquired.

The essence of \eqref{tu} is to maximum  degree of linear association between $\mathbf t$ and $\mathbf u$. Suppose that a linear relation between the latent vectors $\mathbf t$ and $\mathbf u$ exists, \emph{e.i.}  $\mathbf u = \mathbf td +\mathbf h$ , where $d$ is a constant, $\mathbf h$ is error term, and $d$ and $\mathbf h$ can be absorbed by $\mathbf q$ and $\mathbf F$,  respectively.  Based on this relation, \eqref{pls1} can be casted  as the following formula
\begin{equation}\label{pls2}
\mathbf X = \mathbf t\mathbf p^T+\mathbf E,\ \
\mathbf Y = \mathbf t\mathbf q^T+\mathbf F
\end{equation}

 Thus $\mathbf p=\mathbf X^T\mathbf t(\mathbf t^T\mathbf t)^{-1}$ and $\mathbf q=\mathbf Y^T\mathbf t(\mathbf t^T\mathbf t)^{-1}$ can be obtained by the least square method. Then $\mathbf X$ and $\mathbf Y$ can be updated
\begin{equation}
\mathbf X: = \mathbf X-\mathbf t\mathbf p^T,\ \
\mathbf Y: = \mathbf Y-\mathbf t\mathbf q^T
\end{equation}
This procedure is re-iterated $c$ times, and we can obtain the projection matrix $\mathbf W = [\mathbf w_1,\mathbf w_2,\ldots,\mathbf w_c]$, latent components $\mathbf T = [\mathbf t_1,\mathbf t_2,\ldots,\mathbf t_c]$, loading matrices $\mathbf P = [\mathbf p_1,\mathbf p_2,\ldots,\mathbf p_c]$ and $\mathbf Q=[\mathbf q_1,\mathbf q_2,\ldots,\mathbf q_c]$. And \eqref{pls2} can be recast as
\begin{equation}\label{pls3}
\mathbf X = \mathbf T\mathbf P^T+\mathbf E,\ \
\mathbf Y = \mathbf T\mathbf Q^T+\mathbf F
\end{equation}
According to $\mathbf T = \mathbf X\mathbf W$, $\mathbf Y = \mathbf X\mathbf W\mathbf Q^T+\mathbf F$ and regression coefficient $\mathbf R = \mathbf W\mathbf Q^T$.

\section{The PLSR on Riemannian Manifolds}\label{Sec:III}

The core of PLSR is to optimize the squared covariance between latent components $\mathbf T$ and the data $\mathbf Y$, see \eqref{ww}. Boulesterx and Strimmer  \cite{BoulesteixStrimmer2007} had summarized  several different model modification for optimizing the projection matrix $\mathbf W$ in Euclidean spaces. However all the algorithms take a greedy strategy to calculate all the factors one by one, and thus often result in suboptimal solutions. In order to overcome this shortcoming, this paper will take those models as optimization on Riemannian manifolds, and propose an algorithms for solving the projection matrix $\mathbf W$ thus the  latent component matrix $\mathbf T$ as a whole on Riemannian manifolds.

\subsection{ SIMPLSR on  the Generalized Grassmann Manifolds}

We can transform model \eqref{ww} into following optimization problem
\begin{equation}\label{wwi}
    \max_{\mathbf W}  \text{tr}(\mathbf W^T\mathbf X^T\mathbf Y\mathbf Y^T\mathbf X\mathbf W) \  \  \text{s.t.} \ \ \mathbf W^T\mathbf W = \mathbf I
\end{equation}
where $\mathbf W\in\mathbb{R}^{p\times c}$ and  $\mathbf I$ is the identity matrix.

Because of the orthogonal constraint, this constrained optimization problem can be taken as unconstrained optimization on Stiefel manifold
\begin{equation}\label{wws}
    \max_{\mathbf W\in St(p,c)}  \text{tr}(\mathbf W^T\mathbf X^T\mathbf Y\mathbf Y^T\mathbf X\mathbf W).
\end{equation}
\begin{algorithm}[htbp]
\renewcommand{\algorithmicrequire}{\textbf{Input:}}
\renewcommand\algorithmicensure {\textbf{Output:}}
\caption{ SIMPLSR on generalized Grassmann manifold (PLSRGGr)  }\label{algGG}
\begin{algorithmic}[1]
\REQUIRE  matrices $\mathbf X\in \mathbb R^{n\times p},\mathbf Y\in \mathbb R^{n\times q}$.
\STATE Initial  matrix $\mathbf W_1 $ is a randomly generated matrix, gradient norm tolerance $\epsilon_1$, step size tolerance $\epsilon_2$ and  max iteration number $N$. Let  $0<c<1$ $\beta_1$ =0 and $\boldsymbol{\zeta}_0 = \mathbf 0$.
\FOR{$k = 1:N$}
\STATE Compute gradient in Euclidean space
$$\text{grad}_Ef(\mathbf W_k) = 2\mathbf X^T\mathbf Y\mathbf Y^T\mathbf X\mathbf W_k$$
\STATE Compute gradient on generalized Grassmann manifold  $$\boldsymbol{\eta}_k = \mathcal P_{[\mathbf W_k]}(\text{grad}_Ef(\mathbf W_k)),$$ where Projection
operator
\begin{align}
&\mathcal P_{\mathbf W_k}(\mathbf Z) = \mathbf Z- \mathbf W_k\text{symm}(\mathbf W_k^T\mathbf X^T\mathbf X\mathbf Z),\notag\\
&\text{symm}(\mathbf D)=(\mathbf D+\mathbf D^T)/2.\notag
\end{align}
\IF{$k\geq 2$}
\STATE Compute the weighted value\\
$$\beta_k = \text{tr}(\boldsymbol{\eta}_k^T\boldsymbol{\eta}_k)/\text{tr}(\boldsymbol{\eta}_{k-1}^T\boldsymbol{\eta}_{k-1})$$
\STATE Compute a transport direction\\
$$\mathcal T_{\mathbf W_{k-1}\rightarrow \mathbf W_{k}}(\boldsymbol{\zeta_{k-1}})=\mathcal P_{\mathbf W_k}(\boldsymbol{\zeta_{k-1}}).$$
\ENDIF
\STATE Compute a conjugate direction $$\boldsymbol{\zeta}_k = -\text{grad}_Rf(\mathbf W_{k})+\beta_k\mathcal T_{\mathbf W_{k-1}\rightarrow \mathbf W_{k}}(\boldsymbol{\zeta_{k-1}}).$$
\STATE Choose a step size $\alpha_k$ satisfying the Armijo criterion
$$f(R_{\mathbf W_k}(\alpha_k \boldsymbol{\zeta}_k) \geq f(\mathbf W_k) + c\alpha_k \text{tr}(\boldsymbol{\eta}_k^T\boldsymbol{\zeta}_k),$$
where Retraction operator
$$\mathcal R_{[\mathbf W_k]}(\boldsymbol{\zeta}_k)=\mathbf U \mathbf Es\mathbf Ev^T\mathbf V^T;$$
$\mathbf W_k +\boldsymbol{\zeta}_k = \mathbf U\boldsymbol{\Sigma}\mathbf V,$ (SVD decomposition)
$\mathbf U^T\mathbf X^T\mathbf X\mathbf U = \mathbf Ev\mathbf E_{\lambda}\mathbf Ev^{-1}$ (eigendecomposition), $\mathbf Es=\mathbf Ev(\mathbf E_{\lambda})^{-1/2}$.
Set $\mathbf W_{k+1} = R_{\mathbf W_k}(\alpha_k \boldsymbol{\zeta}_k)$.
\STATE  Terminate and output $\mathbf W_{k+1}$ if one of the stopping conditions is satisfied  $\|\boldsymbol{\eta}_{k+1}\|_F\leq \epsilon_1$, $\alpha_k\leq \epsilon_2$ and $k\geq N$
is achieved.
\ENDFOR
\STATE  $\mathbf W = \mathbf W_{k+1}$.
\STATE  Compute $\mathbf T =\mathbf X\mathbf W$.
\STATE  Compute $\mathbf P =\mathbf X^T\mathbf  T(\mathbf  T^T\mathbf T)^{-1}$.
\STATE  Compute $\mathbf Q =\mathbf Y^T\mathbf  T(\mathbf  T^T\mathbf T)^{-1}$.
\STATE  Compute  regression coefficient $\mathbf R = \mathbf W\mathbf Q^T$.
\ENSURE $\mathbf{W, T, P, Q, R}$.
\end{algorithmic}
\end{algorithm}


To represent the data sets $\mathbf X$ and $\mathbf Y$ from \eqref{pls3}, it is more reasonable to constrain latent components $\mathbf T$ in an orthogonal space.
Thus model \eqref{wws} can be rewritten as
\begin{equation}\label{wxxwi}
\begin{split}
   &\max_{\mathbf W} \text{tr}(\mathbf W^T\mathbf X^T\mathbf Y\mathbf Y^T\mathbf X\mathbf W),\\
   &\text{s.t.}  \mathbf T^T\mathbf T=\mathbf W^T\mathbf X^T\mathbf X\mathbf W =\mathbf I.
\end{split}
\end{equation}


Similar to model \eqref{wwi}, we can first convert problem \eqref{wxxwi} to an unconstrained problem on the generalized Stiefel manifold with $\mathbf B = \mathbf X^T\mathbf X$, i.e.,
\begin{equation}\label{wwGS}
    \max_{\mathbf W\in GSt(p,c, \mathbf X^T\mathbf X)}  \text{tr}(\mathbf W^T\mathbf X^T\mathbf Y\mathbf Y^T\mathbf X\mathbf W).
\end{equation}

Let $f(\mathbf W) =  \text{tr}(\mathbf W^T\mathbf X^T\mathbf Y\mathbf Y^T\mathbf X\mathbf W)$ be defined on generalized Stiefel manifold $GSt(p,c,\mathbf B)$. For any  matrix $\mathbf U\in\mathcal{O}(c)$, we have $f(\mathbf W\mathbf U) = f(\mathbf W)$. This means that the maximizer of $f$ is unidentifiable on generalized Stiefel in the sense that if $\mathbf W$ is a solution to \eqref{wwGS}, then so is $\mathbf W\mathbf U$ for any $\mathbf U\in\mathcal{O}(c)$. This may cause some trouble for numerical algorithms for solving \eqref{wwGS}.

 If we contract all the generalized Stiefel points in its equivalent class $[\mathbf W] = \{\mathbf W \mathbf U| \text{ for all } \mathbf U\in\mathcal{O}(c)\}$ together, it is straightforward to convert the optimization \eqref{wwGS} on generalized Stiefel manifold to the generalized Grassmann manifold $GGr( p,c,\mathbf B)$ \cite{EdelmanAriasSmith1998} as follows
 \begin{equation}\label{gGrassprob}
  \max_{[\mathbf W]\in GGr(p,c)} \text{tr}(\mathbf W^T\mathbf X^T\mathbf Y\mathbf Y^T\mathbf X\mathbf W),
\end{equation}
The model \eqref{gGrassprob} is called as statistically inspired modification of PLSR (SIMPLSR) on generalized Grassmann manifolds.

We will use the metric  $g_{[\mathbf W]}(\mathbf Z_1,\mathbf Z_2) = \text{tr}(\mathbf Z_1^T\mathbf B\mathbf Z_2)$ on generalized Grassmann manifold. The matrix representation of the tangent
space of the generalized Grassmann manifold  is identified with a subspace of the tangent space of the total space  that does not produce a displacement along the equivalence classes. This subspace is called the horizontal space \cite{MishraSepulchre2014}. The horizontal
space $ \mathcal H_{[\mathbf W]}GGr(p,c) =\{\mathbf Z\in\mathbb R^{p\times c}: \mathbf W^T\mathbf Z=0\}$.
The other related ingredients such as projection operator, retraction operator, transport operator  for implementing an off-the-shelf nonlinear conjugate-gradient algorithm \cite{TanTsangWangVandereyckenPan2014} for \eqref{gGrassprob} are listed in Algorithm \ref{algGG} which is the optimization algorithm of PLSR on generalized Grassmann manifold.

\subsection{SIMPLSR on Product Manifolds}

Another equivalent expression for SIMPLSR \cite{BoulesteixStrimmer2007} which often appear in the literature is as follows
\begin{align}\label{wxxu}
  &\max_{(\mathbf W,\mathbf U)} \text{tr}(\mathbf W^T\mathbf X^T\mathbf Y\mathbf U),\\
  & \text{s.t.} \ \ \mathbf T^T\mathbf T=\mathbf W^T\mathbf X^T\mathbf X\mathbf W =\mathbf I \ \text{and}\ \ \text{diag}(\mathbf U^T\mathbf U) = \mathbf I. \notag
\end{align}
The feasible domain of $\mathbf W$ and $\mathbf U$ can be considered as a product manifold of a generalized Stiefel manifold $GSt(p,c,\mathbf B)$ with $\mathbf B = \mathbf X^T\mathbf X$  (see \eqref{gst}) and  Oblique manifold $Ob(q,c)$ (see \eqref{obl}), respectively.
The product manifold is denoted as
\begin{equation}
\begin{split}
  &GSt(p,c,\mathbf B)\times Ob(q,c) \\
= &\{(\mathbf W, \mathbf U): \mathbf W\in GSt(p,c,\mathbf B), \mathbf U\in Ob(q,c)\}.
\end{split}
\end{equation}
\begin{algorithm}[hbp]
\renewcommand{\algorithmicrequire}{\textbf{Input:}}
\renewcommand\algorithmicensure {\textbf{Output:}}
\caption{ SIMPLSR on product manifold (PLSRGStO)  }\label{alg4}
\begin{algorithmic}[1]
\REQUIRE  matrices $\mathbf X\in \mathbb R^{n\times p},\mathbf Y\in \mathbb R^{n\times q}$.
\STATE Initial matrices $\mathbf W_1 $ and $\mathbf U_1$ are  randomly generated  matrices, gradient norm tolerance $\epsilon_1$, step size tolerance $\epsilon_2$ and max alternating iterations $N_1$,  max iteration number $N_2$. . Let $0<c<1$ $\beta_1$ =0 and $\boldsymbol{\zeta}_0 = \mathbf 0$.
\FOR{$k = 1:N_1$}
\FOR{$i = 1:N_2$}
\STATE Compute gradient in Euclidean space\\
$\text{grad}_Ef_{\mathbf W}(\mathbf W_i) = \mathbf X^T\mathbf Y\mathbf U_1$
\STATE Some related ingredients of generalized Stiefel manifold are same with generalized Grassmann manifold, and $\mathbf W$ can be solved by Algorithm \ref{algGG}.
\ENDFOR
\STATE $\mathbf W_1 = \mathbf W_i$.
\FOR{$j = 1:N_2$}
\STATE Compute gradient in Euclidean space\\
 $$\text{grad}_Ef_{\mathbf U}(\mathbf U_j) = \mathbf Y^T\mathbf X\mathbf W_1$$
\STATE Compute gradient on Oblique manifold  $$\boldsymbol{\eta}_j = \mathcal P_{\mathbf U_j}(\text{grad}_Ef(\mathbf U_j)),$$  where Projection operator $$\mathcal P_{\mathbf U_j}(\mathbf Z) = \mathbf Z- \mathbf U_j\text{diag}(\mathbf U_j^T\mathbf Z)$$.
\IF{$j\geq 2$}
\STATE Compute the weighted value\\
$$\beta_j = \text{tr}(\boldsymbol{\eta}_j^T\boldsymbol{\eta}_j)/\text{tr}(\boldsymbol{\eta}_{j-1}^T\boldsymbol{\eta}_{j-1})$$
\STATE Compute a transport direction\\
$$\mathcal T_{\mathbf U_{j-1}\rightarrow \mathbf U_{j}}(\boldsymbol{\zeta_{j-1}})=\mathcal P_{\mathbf U_j}(\boldsymbol{\zeta_{j-1}}).$$
\STATE Compute a conjugate direction $$\boldsymbol{\zeta}_j = -\boldsymbol{\eta}_j+\beta_j\mathcal T_{\mathbf U_{j-1}\rightarrow \mathbf U_{j}}(\boldsymbol{\zeta_{j-1}}).$$
\ENDIF
\STATE Choose a step size $\alpha_j$ satisfying the Armijo criterion
\begin{equation*}
f(R_{\mathbf U_j}(\alpha_j \boldsymbol{\zeta}_j) \geq f(\mathbf U_j) + c\alpha_j \text{tr}(\boldsymbol{\eta}_j^T\boldsymbol{\zeta}_j).
\end{equation*}
where Retraction operator
$
\mathcal R_{\mathbf U_j}(\boldsymbol{\zeta}_j)=(\mathbf U_j+\boldsymbol{\zeta}_j)(\text{diag}((\mathbf U_j+\boldsymbol{\zeta}_j)^T(\mathbf U_j+\boldsymbol{\zeta}_j)))^{-1/2} .
$
\STATE  Terminate and output $\mathbf U_{j+1}$ if one of the stopping conditions is satisfied  $\|\boldsymbol{\eta}_{j+1}\|_F\leq \epsilon_1$, $\alpha_j\leq \epsilon_2$ and $j\geq N_2$
is achieved.
\ENDFOR
\STATE $\mathbf U_1 = U_{j+1}$
\ENDFOR
\STATE  Compute $\mathbf T =\mathbf X\mathbf W_1$.
\STATE  Compute $\mathbf P =\mathbf X^T\mathbf  T(\mathbf  T^T\mathbf T)^{-1}$.
\STATE  Compute $\mathbf Q =\mathbf Y^T\mathbf  T(\mathbf  T^T\mathbf T)^{-1}$.
\STATE  Compute  regression coefficient $\mathbf R = \mathbf W\mathbf Q^T$.
\ENSURE $\mathbf{W, T, P, Q, R}$.
\end{algorithmic}
\end{algorithm}

So model \eqref{wxxu} can be modified as
 \begin{equation}\label{wxyu}
   \max_{(\mathbf W, \mathbf U)\in St(p,c,\mathbf B)\times Ob(q,c)} \text{tr}(\mathbf W^T\mathbf X^T\mathbf Y\mathbf U)
\end{equation}
We call this model as equivalent statistically inspired modification of PLSR (ESIMPLSR) on product manifolds.

To induce the geometry of the product manifold,  we use the metric  $g_{\mathbf W}(\mathbf Z_1,\mathbf Z_2) = \text{tr}(\mathbf Z_1^T\mathbf B\mathbf Z_2)$ and the tangent space $ T_{\mathbf W}GSt(p,c,\mathbf B) =\{\mathbf Z\in\mathbb R^{p\times c}: \mathbf W^T\mathbf B\mathbf Z+\mathbf Z^T\mathbf B\mathbf W=\mathbf 0\}$ on the generalized Stiefel manifold, and the metric $g_{\mathbf U}(\mathbf Z_1,\mathbf Z_2) =\text{tr}(\mathbf Z_1^T\mathbf Z_2)$ and the tangent space $ T_{\mathbf U} Ob =\{\mathbf Z\in\mathbb R^{q\times c}: \text{diag}(\mathbf U^T\mathbf Z)=\mathbf 0\}$ on the Oblique manifold.
We optimize model \eqref{wxyu} on the product manifold by alternating directions method (ADM) \cite{BoydParikhChuPeleatoEckstein2011} and nonlinear Riemannian conjugate gradient  method (NRCG), summarized in Algorithm \ref{alg4}. It is the optimization algorithm of PLSR on the generalized Stiefel manifold.

\section{Experimental Results and Analysis}\label{Sec:IV}
In this section, we conduct several experiments on face recognition and object classification on several public databases to assess the proposed algorithms. These experiments are designed to compare the feature extraction performance of the proposed algorithms with existing algorithms including principal component regression (PCR) \cite{NaesMartens1988}
\footnote{PCR and SIMPLS codes are from \url{http://cn.mathworks.com/help/stats/examples.html}}and SIMPLSR \cite{Jong1993}. All algorithms are coded in Matlab (R2014a) and run on a PC machine installed a 64-bit operating system with an intel(R) Core (TM) i7 CPU (3.4GHz with single-thread mode) and 28 GB memory.

In our experiments, face dataset $\mathbf X = [\mathbf X_1, \mathbf X_2,\cdots,\mathbf X_n]$ have $n$ samples from $K$ classes. The $k$th class includes $C_k$ samples. The response data (labels) $\mathbf Y$  can be set as binary matrix,
$$
\mathbf Y_{ik} =
\begin{cases} 1, &\mathbf X_i\in C_k  \\
0,  \ \  &\text{otherwise}.
\end{cases} $$
PLSR are used to estimate the regression coefficient matrix $\mathbf R$
by exploiting training data sets $\mathbf X_{\text{train}}$ and $\mathbf Y_{\text{train}}$. Then the response matrices $\hat{\mathbf Y}_{\text{test}} = \mathbf X_{\text{test}}\mathbf R$ can be predicted for testing data $\mathbf X_{\text{test}}$. We get the  predicted response matrix (predicted labels) $\hat{\mathbf Y}_{\text{test}}$ by setting the largest value to 1 and others to 0 for each row of $\hat{\mathbf Y}_{\text{test}}$ for classification.

\subsection{Face Recognition}
\subsubsection{Data Preparation }

Face data are from the following two public available databases:
\begin{itemize}
  \item The AR face dataset (\url{http://rvl1.ecn.purdue.edu/aleix/aleix face DB.html})
  \item The Yale face dataset (\url{http://www.cad.zju.edu.cn/home/dengcai/Data/Face Data.html})
\end{itemize}

The AR face database consists of over 3,200 frontal color images for 126 people (70 men and 56 women). Each individual has 26 images which were collected in two different sessions separated by two weeks. There are 13 images from each session. 
In experiments, we select data from 100 randomly chosen individuals. The thirteen in first session  of each individual are used for training and the other thirteen in second session for testing. Each image is cropped and resized to 60 $\times$ 43 pixels, then vectorized as a 2580-dimension vector.

The Yale face database  contains 165 images from 15 individuals. Each individual provides  11 different images. 
In the experiment, 6 images from each individual are randomly selected as training sample while the remaining images are for testing.  
Each images are scaled to a resolution of $64\times 64$ pixels, then vectorized as a 4096-dimensional vector. 

\subsubsection{Recognition Performance}
we compare the recognition performance of PCR, SIMPLSR,  PLSRGGr and PLSGRStO on both AR and Yale face datasets.
\begin{figure}[htb]
\begin{center}
{\includegraphics[width=0.45\textwidth]{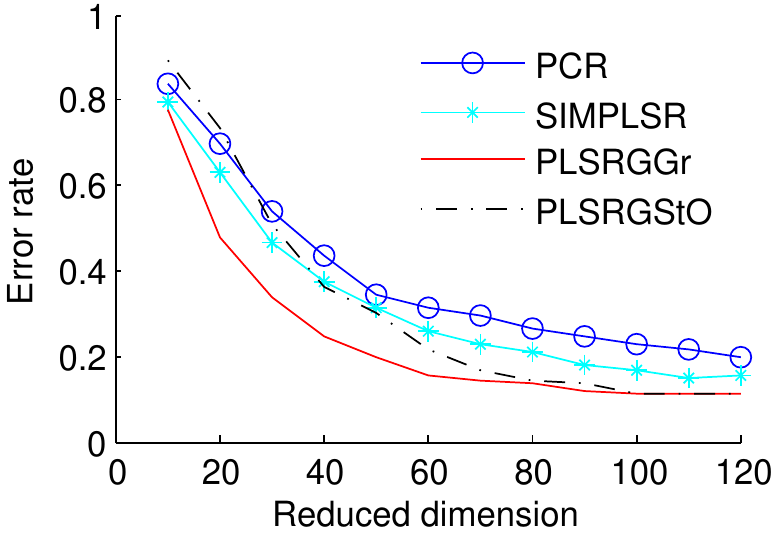}}
\end{center}
\caption{Recognition error (\%)  on AR face database.}\label{AR26}
\end{figure}

Figure \ref{AR26} reports the experiment results on AR face database. It shows that the recognition performance of our proposed algorithms, PLSRGGr and PLSRGStO, is better than other methods more than 4 percent when reduced dimension is greater than 60. Obviously, PLSRGGr has good performance all the time. This demonstrates that our proposed optimization models and algorithms of PLSR on Riemannian manifold  significantly enhances the accuracy. The reason is that  calculating PLSR factors as a whole on Riemannian manifolds can obtain the optimal solution.

\begin{table*}[htb]
  \centering
   \begin{tabular}{|c|c|c|c|c|c|}
   \hline
    $c$  &PCR  &SIMPLSR&PLSRGGr&PLSRGStO  \\
  \hline
      12& 33.27$\pm$3.10    & 26.00$\pm$4.59    &$\mathbf{12.00\pm0}$   & 20.07$\pm$0.30\\
  \hline
      13& 30.87$\pm$4.66      & 21.73.00$\pm$4.06    &$\mathbf{9.40\pm0.30}$   & 15.93$\pm$0.30\\
  \hline
     14& 29.27$\pm$4.79      & 19.00$\pm$3.82         &$\mathbf{8.00\pm0}$   & 8.13$\pm$0.60\\
\hline
        15& 25.87$\pm$4.61      & 15.80$\pm$3.13         &$\mathbf{10.60\pm0.30}$   & 10.67$\pm$0\\
  \hline
\end{tabular}
    \caption{Recognition error (\%)  on Yale face database. }\label{Yaletable}
\end{table*}
\begin{table*}
  \centering
   \begin{tabular}{|c|c|c|c|c|c|c|c|c|}
  \hline
     $c$&GDA&DCC&LSRM&PCR  &SIMPLSR&PLSRGGr&PLSRGStO  \\
        \hline
           5&- &-&-& 23.75 &26.25  &$\mathbf{18.75}$ & $26.25$    \\
  \hline
           6&- &-&-&  22.50 &15.00 &$15.00$ & $\mathbf{13.75}$    \\
  \hline
           7& - &-&-& 21.25 &7.50 &$\mathbf{1.25}$ & $\mathbf{1.25}$    \\
  \hline
           8& 2.50 &11.20&5.00& 20.00 &3.75  & $\mathbf{1.25}$ & $\mathbf{1.25}$    \\
  \hline
\end{tabular}
    \caption{Classification error (\%)  on ETH-80 database, the error rate in last line is employed for GDA, DCC, LSRM.}\label{ETH-80table}
\end{table*}

Another experiment was conducted on Yale face database. In this experiment, the compared algorithms are PCR, SIMPLSR, PLSRGGr and PLSRGStO, and every algorithm is run 20 times. Table \ref{Yaletable} lists the recognition error rates including their mean and standard deviation values with reduced dimensions $c = 12, 13, 14, 15$.  From the table we can observe that the mean  of recognition error rates of PLSRGGr and PLSRGStO is superior to others with a margin of 5 to 14 percentages, and the standard deviation is also smaller.  This demonstrates that our proposed methods more robust. The bold figures in the table highlight the best results for comparison.

\subsection{Object Classification }
\subsubsection{Data Preparation }
For the object classification tasks, we use the following two public available databases for testing,
\begin{itemize}
  \item COIL-20 dataset (\url{http://www.cs.columbia.edu/CAVE/software/softlib/coil-20.php});
  \item ETH-80 dataset  (\url{http://www.mis.informatik.tu-darmstadt.de/Research/Projects/categorization/eth80-db.html}).
\end{itemize}

 Columbia Object Image Library (COIL-20) contains 1,440 gray-scale images from 20 objects. Each object offers 72 images. 
 36 images of each object were selected by equal interval sampling as training while the remaining images are for testing. 

 ETH-80 database \cite{LeibeSchiele2003} consists of 8 categories of 
 objects
 Each category contains 10 objects with 41 views per object, spaced equally over the viewing hemisphere, for a total of 3280 images. 
 Images are resized to $32\times32$ pixels with grayscale pixels and vectorized as 1024-dimensional vector. For each category and each object, we model the pose variations by a subspace of the size $m =7$, spanned by the 7 largest eigenvectors from SVD. In our experiments, the Grassmann distance measure between two point $\text{span}(\mathbf X),\text{span}(\mathbf Y) \in Gr(n,m),$  is defined as
$ \text{dist}(\mathbf X,\mathbf Y) = \| \arccos(\text{svd}(\mathbf X^T\mathbf Y))\|_F$
 which is the F-norm of principal angles \cite{WolfShashua2003},
 $\text{svd}(\mathbf X^T\mathbf Y)$ denotes the singular value of $\mathbf X^T\mathbf Y$. We follow the experimental protocol from \cite{HammLee2008} which is tenfold cross validation for image-set matching.

 \subsubsection{Classification Performance}
 \begin{figure}[htb]
   \begin{center}
{\includegraphics[width=0.45\textwidth]{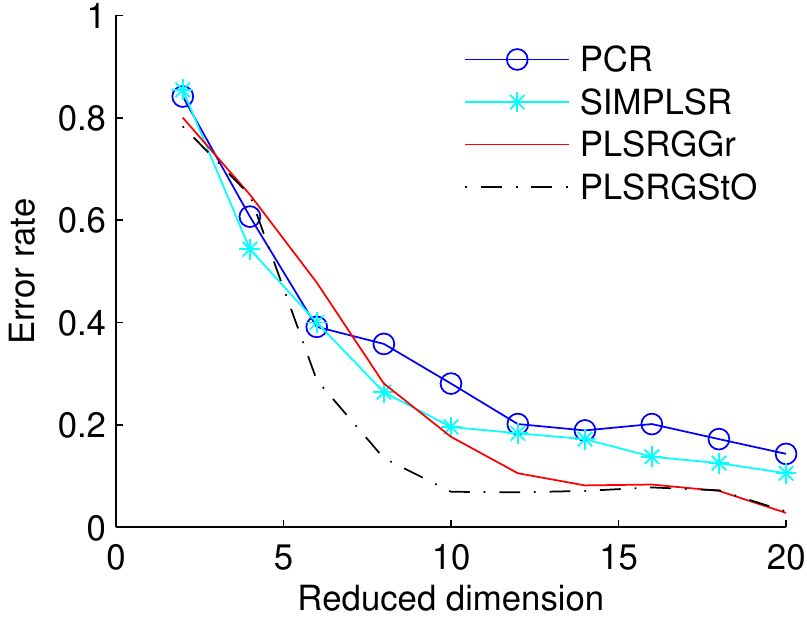}}
    \end{center}
    \caption{Classification error (\%)  on COIL20 database. }\label{Coilf}
\end{figure}
 Figure  \ref{Coilf} lists the classification error of four algorithms on COIL-20 database. The classification errors are recorded for the different reduced dimension $c =\{17, 18,19,20\}$, respectively. From the results,  it can  be found that the proposed methods, PLSRGGr and PLSRGStO,  outperform their compared non-manifold methods with a margin of 2 to 10 percentages when reduced dimension is greater than 10.


To demonstrate the effectiveness of our regression algorithms on the ETH-80 data set. We compared with several contrast methods. Table \ref{ETH-80table} reports the experimental results  with reduced dimension $c = \{5, 6, 7 , 8\}$. The results of  GDA (Grassmann discriminant analysis) \cite{HammLee2008},  DCC (Discriminant canonical correlation) \cite{KimKittlerCipolla2007},  LSRM (Least squares regression on manifold) \cite{Lui2016}  in last line of Table \ref{ETH-80table} are from \cite{Lui2016}. Compared with state of-the-art algorithms, our proposed methods, PLSRGGr and PLSRGStO, both outperform all of them.

\section{Conclusions}\label{Sec:V}
In this paper, we developed  PLSR optimization  models on both Riemannian manifolds, \emph{i.e.} generalized Grassmann manifold and product manifold. We also gave optimization algorithms on both the Riemannian manifolds, respectively. Each of new  models transforms the corresponding original constrained optimization problem to an unconstraint optimization on Riemannian manifolds. This makes it possible  to calculate all the  PLSR factors  as a whole to obtain the optimal solution. The experimental results show our proposed PLSRGGr and PLSRGStO outperform other  methods on several public datasets.
\begin{quote}
\begin{small}
\bibliographystyle{aaai}
\bibliography{reference_haoran}
\end{small}
\end{quote}

\end{document}